\documentclass[conference,a4paper]{IEEEtran}

\makeatletter
\def\endthebibliography{%
	\def\@noitemerr{\@latex@warning{Empty `thebibliography' environment}}%
	\endlist
}
\makeatother

\IEEEoverridecommandlockouts
\usepackage{cite}
\usepackage{amsmath,amssymb,amsfonts}
\usepackage{algorithmic}
\usepackage{graphicx}
\usepackage{textcomp}
\usepackage{xcolor}
\usepackage{subfig}
\usepackage{hyperref}
\def\BibTeX{{\rm B\kern-.05em{\sc i\kern-.025em b}\kern-.08em
    T\kern-.1667em\lower.7ex\hbox{E}\kern-.125emX}}
\begin{document}

\title{Semantic-Aware Environment Perception for Mobile Human-Robot Interaction\\
\thanks{This work was supported by the Federal Ministry of Education and Research of Germany (BMBF) (Project RoboAssist) under Grant 03ZZ0448L.}
}
\author{\IEEEauthorblockN{1\textsuperscript{st} Hempel, Thorsten}
\IEEEauthorblockA{\textit{EIT} \\
\textit{Otto-von-Guericke-University}\\
Magdeburg, Germany \\
thorsten.hempel@ovgu.de}
\and
\IEEEauthorblockN{2\textsuperscript{nd} Fiedler, Marc-André}
\IEEEauthorblockA{\textit{EIT} \\
\textit{Otto-von-Guericke-University}\\
Magdeburg, Germany \\
marc-andre.fiedler@ovgu.de}
\and
\IEEEauthorblockN{3\textsuperscript{rd} Khalifa, Aly}
\IEEEauthorblockA{\textit{EIT} \\
	\textit{Otto-von-Guericke-University}\\
	Magdeburg, Germany \\
	aly.khalifa@ovgu.de}
\and
\IEEEauthorblockN{4\textsuperscript{th} Al-Hamadi, Ayoub}
\IEEEauthorblockA{\textit{EIT} \\
	\textit{Otto-von-Guericke-University}\\
Magdeburg, Germany \\
ayoub.al-hamadi@ovgu.de}
\and
\IEEEauthorblockN{5\textsuperscript{th} Dinges, Laslo}
\IEEEauthorblockA{\textit{EIT} \\
	\textit{Otto-von-Guericke-University}\\
	Magdeburg, Germany \\
	laslo.dinges@ovgu.de}

}

\maketitle
\begin{abstract}
Current technological advances open up new opportunities for bringing human-machine interaction to an new level of human-centered cooperation. In this context, a key issue is the semantic understanding of the environment in order to enable mobile robots more complex interactions and a facilitated communication with humans. Prerequisites are the vision-based registration of semantic objects and humans where the latter are further analyzed for potential interaction partners. Despite significant research achievements, the reliable and fast registration of semantic information still remains a challenging tasks for mobile robots in real-world scenarios. 
In this paper, we present a vision-based system for mobile assistive robots to enable a semantic-aware environment perception without additional a-priori knowledge. We deploy our system on a mobile humanoid robot that enables us to test our methods in real-world applications.

\end{abstract}

\begin{IEEEkeywords}
human-robot interaction, semantic objects, interaction willingness

\end{IEEEkeywords}

\section{Introduction}
Recently, assistive robotics became an increasingly popular research field as it provides use-case applications for solutions from a wide range of disciplines. Whereas special-purposed service robots such as autonomous cleaning systems or lawn mowers already found their way into many households, more general-task directed robotic assistants still entail challenges for the safe and reliable use in unconstrained environments. In general, these systems are aimed at providing various services in the household and workspace of humans that are assigned in direct and human-oriented interactions. This requires the assistive robot to enhance the mostly geometric perception as solely needed for cleaning tasks to a more general semantic understanding of its environment. In particular, a common task for a service robot is to find and bring back a specific kind of object. In order to execute such  a command the robot must be able to interact with the human by understanding its needs and subsequently having the capabilities to navigate and find the dedicated object autonomously.

\begin{figure}[t]
	\centering
	\includegraphics[width=\linewidth]{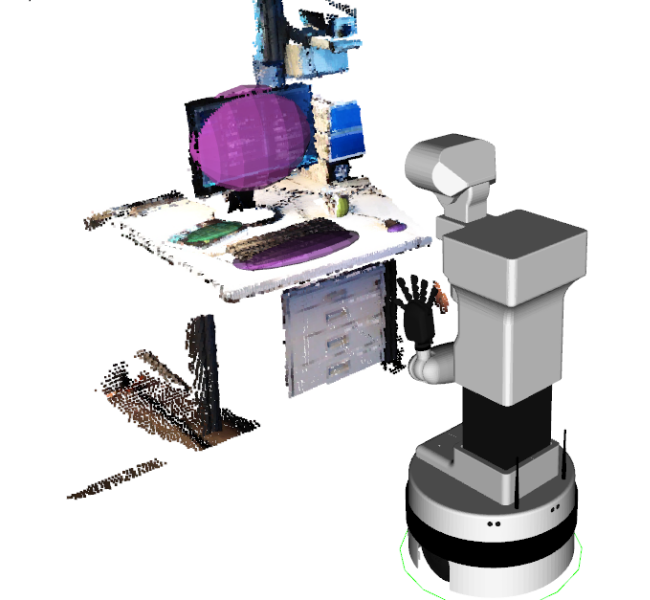}
	\caption{Example image of the TIAGo robot in front of a mapped office environment. Multiple objects on the desk, such as screen, cup, book, mouse and keyboard, are additionally incorporated as projected spheres into the point cloud map.}
		\label{info}
\end{figure}
In this paper we focus on the perceptual semantics for processing meaning-oriented information for objects and humans. 
Multiple approaches \cite{hobbit, martinez, lee} have been presented targeting the detection and recognition of semantic data for mobile robots. Common methods harness model-based solution approaches where each object is identified by its shape or color using handcrafted descriptors. Given that the target object has an unique appearance, these methods can be very performant for recognition tasks. On the other side, low image resolution, small objects, an ordinary appearance and multiple instances of similar looking objects can impair the detection and recognition of such systems. In addition, the target models have to be taught beforehand with training images of the object showing it from multiple different perspectives.

We tackle this challenge by using neural networks and combine them with geometric constraints provided by the robot. Our system can be divided in two separate modules. The first one targets the enrichment of geometric maps with semantic information by detecting and localizing objects in a robust manner (Fig. \ref{info}). Additionally, we make sure to constantly correct and extend currently mapped objects based on updated perception data from the robot. This makes our system also suitable for localization methods that provide retrospective optimization capabilities. The procedure is carried out on the fly in real-time, when the robot is exploring the surroundings making it a well benefiting add-on for geometric-based mapping algorithms.  
The second module detects and analyzes potential human interaction partners where we aim to additionally predict cooperation willingness. This provides a first step of interactions with humans in a proactive manner while at the same time semantic objects can be adhoc incorporated in resulting tasks.
We implemented our system on the TIAGo, a mobile humanoid robot platform.
The rest of this paper is organized as follows: Section \ref{sec:system} describes our proposed system. In section \ref{implementation}, we give details of the current state of implementation. Section \ref{conclusion} concludes and gives an insight about our future work.
\begin{figure*}[th]
	\centering
	\includegraphics[width=\linewidth]{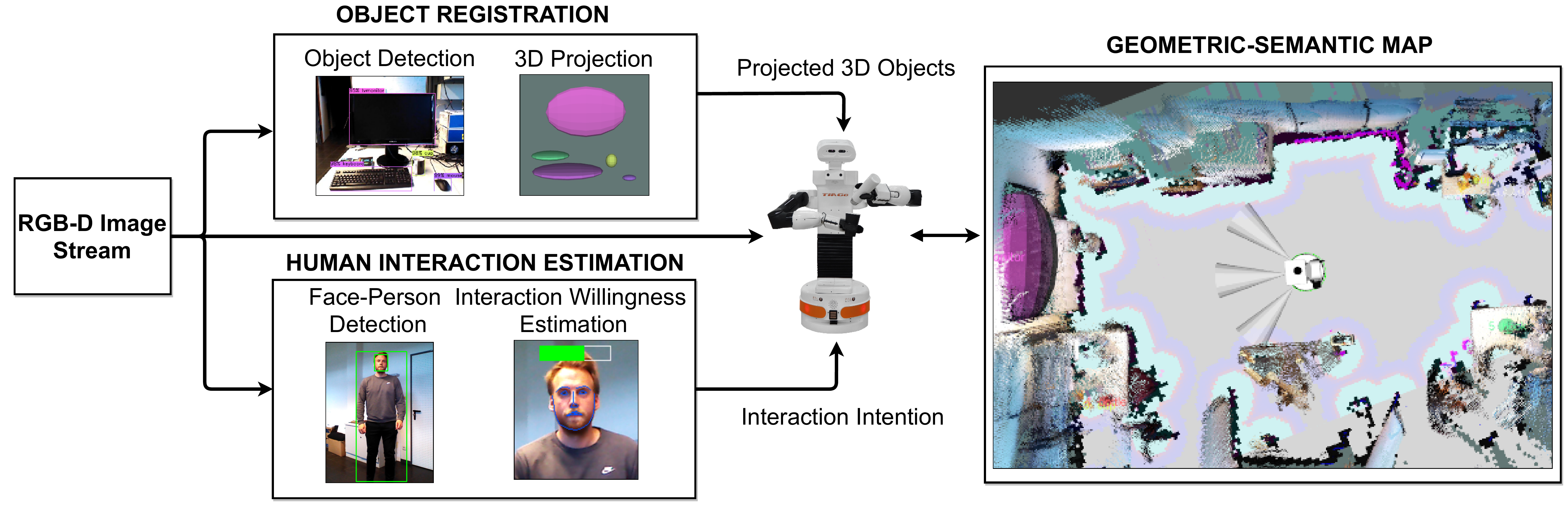}

	\caption{Overview of the proposed method. We use RGB-D images to feed our process lines for object registration and human condition estimation. The object registration consist of 2D object detections that are mapped and associated to 3D objects. The human condition estimation starts with multi task detection process for prediction persons and faces that are in a subsequent step further analyzed for interaction willingness. }
	\label{system}
\end{figure*}

\section{system components} 
\label{sec:system}
Our system (Fig. \ref{system}) consists of two main pipelines for extracting and evaluating semantic information from visual data. The first one is dedicated to detect and register volumetric objects. The second one classifies and estimates the attention related behavior of humans. In the following subsections, we describe each component in detail.

\subsection{Object registration}       
The first step of the object registration is the classification of image regions in the image stream provided by the camera in the head of the robot. For this matter deep neural networks have been proven to reliably detect wide ranges of different types of objects. Popular architectures are the RCNN model family \cite{rcnn, FasterRCNN} and the YOLO model family \cite{yolo, yolo9000}. Due to it outstanding speed we chose YOLOv4 \cite{yolov4} for our framework. The network is pretrained on the COCO database\cite{coco} and able to predict up to 80 different classes. However, when testing it outside of the database we experienced a quite decreased accuracy rate and repeating occurrences of false positive detections. Accordingly, the unfiltered processing of the provided predictions will result with false labeled map parts in later stages. To overcome this problem, we apply an adapted Intersection over Union (IOU) tracker\cite{iou} that associates similarly located and equally labeled bounding boxes from consecutive images. Associated detections form a so called \textit{track}, where the confidence of a true positive object classification increases with the tracks length.
After reaching a specific track length threshold the 2D detection is projected into the 3D space using the intrinsic camera parameters and the extrinsics provided by the robots localization algorithm. For this purpose the depth estimation from the camera active stereo module is used together with the RGB images to generate metric scaled point clouds. Depending on the bounding box sizes we then cut out object cuboids that are afterwards projected into the robots 3D point cloud map. While this method for registering of semantic objects can be performed in a very efficient way, it is combined with supportive solutions to overcome the following challenges: 
 \linebreak
\subsubsection{Object Recognition}
When the robot is moving around it may detect objects in the image stream that already has been registered earlier to the map. Processing it a second time will lead to multiple mapped instances of the same physical objects and must be therefore prevented. Consequently, the object mapping must be able to recognize objects that has been seen and registered before. In general, 3D object recognition has been extensively researched in the recent time, but it commonly requires huge computational efforts. In addition, objects from the same class can resemble each other in a way that taking unique visual fingerprints fails. We approach the issue by comparing the projected localization of a new object \textit{candidate} with the localization of earlier registered objects of the same class. Is the candidate significantly overlaying with its counterpart it is likely that both belong to the same origin. This approach is implemented using a \textit{nearest neighbor} association process, where we compare the average distance between the entire point clouds instead of their centroids. This way, we incorporate the objects size in the association decision as point clouds of large objects can contain centroids that are wide apart, while the clouds themselves heavily overlap. 
 \linebreak
\subsubsection{Object Optimization}
The probabilistic localization of the robot involves uncertainties and will only be a convergence of its real pose. This effect is further enhanced when navigating in an unknown environment and, thus, a simultaneous mapping process is required. For localization methods with optimization capabilities the detection of distinctive or familiar map areas is exploited to re-evaluate the past trajectory and, where appropriate, to correct drifts. Concurrently, we use this mechanism to recalculate the poses of our semantic objects derived from the corrected robot trajectory. At the same time, we analyze if this leads to any strong overlaps between their updated point clouds. This provides us the opportunity to even correct mapping errors induced from the erroneous trajectory. Objects with salient overlap are assumed to belong to the same physical instance and are therefore merged whereby the points clouds are concatenated and the localization and object dimensions are recalculated accordingly. This way, we can not only maintain and correct the perception of semantic objects around us, but with the aid of point cloud merging we are able to build up more complete point cloud appearance models of our objects.
\subsection{Human Behavior Estimation}
To create optimal preconditions for a successful human-robot interaction we apply a dedicated process pipeline to search for and analyze human interaction partners. Instead of using one network for person detection and one for face detection, we utilize a custom neural network that is able to predict person and face bounding boxes simultaneously.
 \linebreak 
\subsubsection{Face-Person Detection}
The network architecture is based on the SSD approach~\cite{ssd} that provides a higher inference rate than two-stages detectors. Moreover, both detectors for persons and faces are combined using multi-task learning which enables them to share their first layers of feature extraction. This equally boost the efficiency and the generalization capabilities.  

The main difficulty for the combination of the two tasks face and person detection in a single neural network is the fact that publicly available databases contain only ground truths for one of the two tasks. For this purpose, we developed a custom multi-task loss function and designed an architecture consisting of a shared backbone and separate detection layers for each detection task. During training, we alternate between batches of person annotations and batches of face annotations, which are taken from different databases.
The prediction of the respective class with non-existing ground truth is assumed to be correctly determined and only the gradients of the detection layers with existing ground truth information are adjusted. Thus, a completely end-to-end trainable framework could be created.
  
While the predicted bounding boxes for persons give us a good estimation of its localization, the face predictions are further used for behavior examination. In particular, we want to know if the human is interested in an interaction with the robot. A good indication for general interest is the gaze or head pose pointing towards the robot.
As the gaze estimation is error-prone at low image resolutions, we focus on the head pose to evaluate general interaction willingness.
 \linebreak
\subsubsection{Interaction Willingness Estimation}
To estimate the head pose we first determine the position of facial landmarks in the face images provided by the face predictor. These facial keypoints are distinctive spots in the face (e.g. the corners of the eyes, the sides and top of the nose) that provide us information about the current formation of the face. We estimate the landmark positions by applying an ensemble of regression trees~\cite{faciallm} that has proven to provide very fast and reliable predictions. In the following step we project these 2D landmarks into the 3D space. This is achieved by using a default 3D model that contains the same facial landmarks and aligning it with the previously predicted 2D counterpart. As this represents a minimization problem we apply the commonly used Levenberg-Marquardt optimization to estimate a matching 3D mask. The rotation and translation for the projection that provides the smallest re-projection error implies the head pose.

The prediction is performed for every single image that contains a detected face. However, the pose information from a single frame is not sufficient to derive assumptions about an underlying general interaction willingness. Brief views in the direction of the robot can be caused by arbitrary intentions including behavior estimation (e.g. when the robot is moving) or even causal glances without deeper conscious intentions. We therefore track a humans head pose over multiple images and gain confidence about interaction intentions the longer the persons focus is directed at the robot. We assume that a viewing direction towards the robot of a duration of 3~ms represents a reliable threshold to determine interaction willingness. However, short distractions are common that result in brief interruptions of the attention towards the robot. We therefore use a solution based on our previous idea \cite{hempel} and calculate the interaction willingness in a dynamic manner. We apply a progress bar that loads faster when the attention is directed to the robot and unloads slower in case of distractions. In this way, showing interaction willingness can be resumed in natural way even though it has been abandoned for a short time.   

\section{Implementation}
\label{implementation}
We implemented our system in form of multiple Robot Operating System (ROS) compatible modules for seamless intercommunication with other (ROS) components and deployed it on the TIAGo robot. For mapping and localization we use the RTAB-Map~\cite{rtabmap} as its graph-based approach suits our semantic data update and correction process. The robots IMU is used as guess to perform a laser-based ICP-SLAM that runs along with other ROS nodes for path planning, collision avoidance and motion planning on the robots onboard i7 computer. Our image processing focused methods are deployed on an external mobile system that is placed on the robots shoulders. It contains a Quadro RTX 5000 that is able to process the neural networks more efficiently than CPUs.

\section{Conclusion}
\label{conclusion}
In this work, we addressed the problem of semantic meaningful perception for mobile assistive robots which constitutes a fundamental requirement for solving complex tasks. 

We propose a neural network enhanced approach for successively mapping and maintaining 3D objects of the robots environment that can run alongside other geometrical mapping modules. 
Similarly, we process humans by utilizing a custom single-shot detector that simultaneously provides person and face predictions in the image stream. The latter are furtherly used to estimates the persons interaction willingness to enable proactive collaboration behavior on the robots side. All modules are implemented on a mobile humanoid robot platform and are compatible for intercommunication with other ROS modules such as path and motion planning.    
In future works we will exploit this advantage to incorporate additional text-to-speech and speech recognition modules. First, we will search for potential interaction partners and proactively call for tasks when interaction willingness is predicted. Afterwards the tasks can be provided orally, processed and executed. Typical tasks will be the localization and bringing of specific objects that have either already been mapped or have to be found in a dedicated exploration drive.
The behavior will be evaluated in real world scenarios.  

\bibliographystyle{IEEEtranDOI}
\bibliography{IEEEabrv,IEEEexample}

\vspace{12pt}

\end{document}